\documentclass{article} % For LaTeX2e
\usepackage{iclr2020_conference,times}

\usepackage{amsmath,amsfonts,bm}

\def\eqref#1{equation~\ref{#1}}
\def\1{\bm{1}}

\DeclareMathAlphabet{\mathsfit}{\encodingdefault}{\sfdefault}{m}{sl}
\SetMathAlphabet{\mathsfit}{bold}{\encodingdefault}{\sfdefault}{bx}{n}

\usepackage{graphicx}
\usepackage{hyperref}
\usepackage{url}
\usepackage{bbm}
\usepackage{booktabs}
\usepackage{floatrow}
\usepackage{multirow}
\usepackage{wrapfig}

\newfloatcommand{capbtabbox}{table}[][0.5\textwidth]

\title{Differentiable Reasoning over a \\Virtual Knowledge Base}

\author{Bhuwan Dhingra$^1$%
\thanks{Part of this work was done during an internship at Google.}
, Manzil Zaheer$^2$, Vidhisha Balachandran$^1$,\\
\textbf{Graham Neubig$^1$, Ruslan Salakhutdinov$^1$, William W. Cohen$^2$}\\
$^1$ School of Computer Science, Carnegie Mellon University\\
$^2$ Google Research\\
\texttt{\{bdhingra, vbalacha, gneubig, rsalakhu\}@cs.cmu.edu}\\
\texttt{\{manzilzaheer, wcohen\}@google.com}
}

\iclrfinalcopy % Uncomment for camera-ready version, but NOT for submission.
\begin{document}

\maketitle

\begin{abstract}

We consider the task of answering complex multi-hop questions using a corpus as a \textit{virtual knowledge base (KB)}. 
In particular, we describe a neural module, DrKIT, that traverses textual data like a KB,
softly following paths of relations between
\textit{mentions} of entities in the corpus.
At each step the module
uses a combination of sparse-matrix TFIDF indices and a maximum inner product search (MIPS) on a special index of
contextual representations of the mentions.
% map a set of entities $X$ to all entities $Y$ related to something in $X$ (by some specified relations), as witnessed by some document in the corpus.  To answer multi-hop questions, the set of output entities $Y$ can be again used recursively as the input to a second copy of the module.
This module is differentiable, so the full system can be trained end-to-end using gradient based methods,
starting from natural language inputs.
%Thus, we name it DrKIT: Differentiable Reasoning over a virtual Knowledge base of Indexed Text.
We also describe a pretraining scheme for the contextual representation
encoder by generating hard negative examples using existing knowledge bases.
We show that DrKIT improves accuracy by $9$ points on 3-hop questions in
the MetaQA dataset, cutting the gap between text-based and KB-based state-of-the-art by $70\%$.
On HotpotQA, DrKIT leads to a $10\%$ improvement over a BERT-based re-ranking approach
to retrieving the 
relevant passages
required to answer a question.
DrKIT is also very efficient, processing $10$-$100$x
more queries per second than existing multi-hop
systems.\footnote{%
Code available at
\url{http://www.cs.cmu.edu/~bdhingra/pages/drkit.html}
}
\end{abstract}

\section{Introduction}

Large knowledge bases (KBs), such as Freebase and WikiData,
organize information around entities, which makes it
easy to reason over their contents.
For example, given a query like
\textit{``When was the Grateful Dead's lead singer born?''},
one can
identify the entity \textit{Grateful Dead} and the path of relations
\textit{LeadSinger}, \textit{BirthDate} to efficiently extract the answer---provided that this information is present in the KB.  Unfortunately, KBs are often incomplete \citep{min2013distant}. 
While relation extraction
methods can be used to populate KBs,  this process
is inherently error-prone, expensive and slow.
% and errors in extraction can
% propagate to downstream tasks.
% favors precision over recall \gn{I don't know if this is \emph{necessarily} the case. Maybe just write ``this process is also inherently error-prone, and errors in extraction can propagate to the down-stream QA task''}, resulting in low coverage
% for rare entities and relations.

Advances in open-domain QA \citep{moldovan-etal-2002-performance,
yang2019end} suggest an alternative---instead of
performing relation extraction, one could treat a large corpus
as a virtual KB by answering queries with spans from the corpus.
% In
% other words, given a query, one could retrieve a passage that contains
% the answer and then extract the necessary information on-the-fly,
% without ever explicitly creating a KB.
This ensures facts are not
lost in the relation extraction process, but also poses challenges.
One challenge is that it is relatively expensive to answer questions using QA models which
encode each document in a query-dependent fashion
\citep{chen2017reading,
devlin2018bert}---even with modern hardware
\citep{strubell2019energy,schwartz2019green}.  The cost of QA is especially problematic
for certain complex questions, such as the example question above.  If the passages stating that \textit{``Jerry Garcia was the lead singer of the Grateful Dead''} and \textit{``Jerry Garcia was
born in 1942''} are far apart in the corpus, it is difficult
for systems that retrieve and read a single passage to find an answer---even though in this example, it might be easy to answer the question after the relations were explicitly extracted into a KB.  
More generally, complex questions involving
sets of entities or paths of relations may require aggregating information from 
multiple documents, which is expensive.
% for typical neural QA systems.

One step towards efficient QA is the recent work of
\citet{seo2018phrase,seo2019real} on phrase-indexed question
answering (PIQA), in which spans in the text corpus are associated
with question-independent contextual representations and then {indexed
for fast retrieval}.  Natural language questions are then answered by
converting them into vectors that are used to 
perform
maximum inner
product search (MIPS) against the index.
This can be done efficiently using
approximate algorithms
\citep{shrivastava2014asymmetric}.
% This ensures
% efficiency during inference. 
However, this approach {cannot} be directly used to answer complex queries,
since by construction, the information stored in the index is about the
local context around a span---it can only be used for questions where the answer can be derived by reading a single passage.  

This paper addresses this limitation of phrase-indexed question answering.  We 
introduce an efficient, end-to-end differentiable
framework for doing complex QA over a large text corpus that has been
encoded in a query-independent manner.  
Specifically, we consider ``multi-hop'' complex queries
which can be answered by repeatedly executing a ``soft'' version of the operation below,
defined over a set of entities $X$ and a relation $R$:
\begin{equation*}
    Y = X.\text{follow}(R) = \{x': \exists x \in X \enskip \text{s.t.} \enskip R(x,x') \enskip \text{holds}\}
\end{equation*}
In past work soft, differentiable versions of this operation were used to answer multi-hop questions against an explicit KB \citep{cohen2019neural}.
Here we propose a more powerful neural module which approximates this operation against an indexed corpus (a virtual KB).
% ---i.e., 
% %\vbc{I think some background needed before introducing this? Seems a shift in topic.}
% a \textit{differentiable} %, \textit{parameterized}
% module for executing a single step of a complex query against a virtual KB.
% %which can be thought of as a 
% \gn{Made ``follow'' text, as it's a function call.}
In our module, the input $X$ is a sparse-vector representing a
weighted set of entities, and the relation $R$ is a dense feature vector,
e.g.~a vector derived from a neural network over a natural language query.
$X$ and $R$ are used to construct a MIPS
query used for retrieving
the top-$K$ spans from
the index.
The output $Y$ is another sparse-vector representing the weighted set
of entities, aggregated over entity mentions in the top-$K$ spans.
We discuss pretraining schemes for the index in \S\ref{sec:pretraining}.

For multi-hop queries, the output entities $Y$ can be recursively passed as input to
the next iteration of the same module.  The weights of the entities in $Y$
are differentiable w.r.t~the MIPS queries,
which allows end-to-end learning 
\textit{without} any intermediate supervision.
We discuss an implementation based
on sparse-matrix-vector products,
whose runtime and memory depend \textit{only}
on the number of spans $K$ retrieved from the index.
This is crucial for scaling up to large corpora,
providing up to $15$x faster inference than existing
state-of-the-art multi-hop and open-domain QA systems.
% This leads to fast learning and inference,
% and importantly, 
% allowing us to process more than $15$
% $3$-hop queries per second on a single $6$-core CPU 
% \bd{add some impressive statistics here.}
The system we introduce is called DrKIT (for Differentiable Reasoning over a Knowledge base of Indexed Text).
We test DrKIT on the MetaQA benchmark for
complex question answering,
and show that it improves on prior text-based systems
by $5$ points on $2$-hop and $9$ 
points on $3$-hop questions,
reducing the gap between text-based and KB-based
systems by $30\%$ and $70\%$, respectively.
We also test DrKIT on a new dataset
of multi-hop slot-filling over Wikipedia
articles,
and show that it outperforms
DrQA \citep{chen2017reading}
and PIQA \citep{seo2019real}
adapted to this task.
Finally,
we apply DrKIT to multi-hop information
retrieval on the HotpotQA dataset
\citep{yang2018hotpotqa},
and show that it significantly improves
over a BERT-based reranking approach, while
being  $10$x faster.

\begin{figure}[t]
    \centering
    \includegraphics[width=0.9\textwidth]{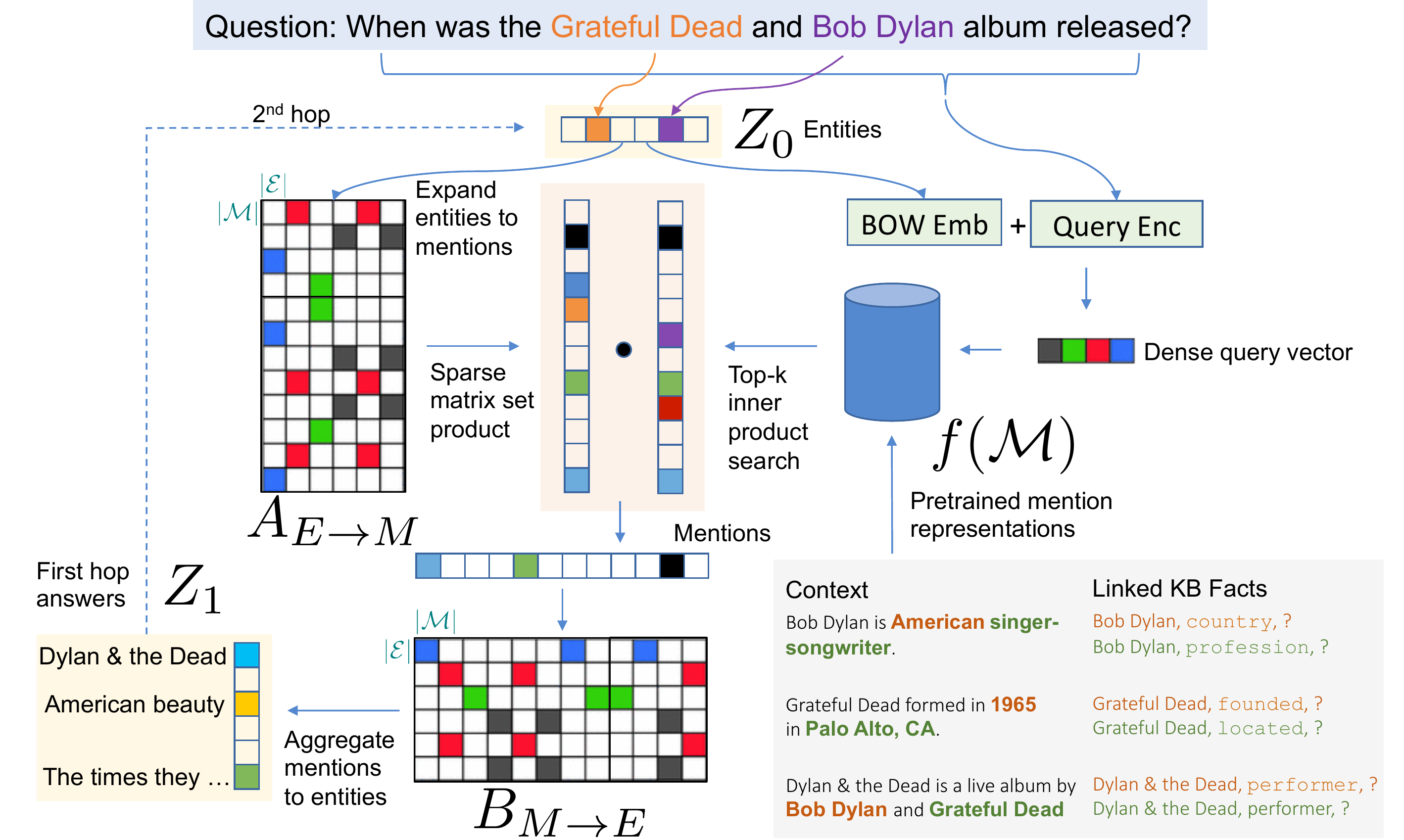}
    \caption{\small DrKIT answers multi-hop questions by iteratively mapping an input set of entities $X$
    (\textit{The Grateful Dead}, \textit{Bob Dylan}) to an output set of entities $Y$ 
    (\textit{Dylan \& the Dead}, \textit{American beauty}, ...) which are related to any input entity by some
    relation $R$ (\textit{album by}).}
    \label{fig:my_label}
\end{figure}

\section{Differentiable Reasoning over a KB of Indexed Text}

We want to answer a question $q$ using a text corpus as if it were a KB. 
We start with the set of entities $z$ in the question $q$, and would ideally want to follow relevant outgoing relation edges in the KB to arrive at the answer. 
To simulate this behaviour on text, we first expand $z$ to set of co-occurring mentions $m$ (say using TFIDF). 
Not all of these co-occurring mentions are relevant for the question $q$,
%so we need to select and retain only relevant ones. 
so we train a neural network which filters
the mentions based on a relevance score of $q$ to $m$. 
Then we can aggregate the resulting set of mentions
$m$ to the entities they refer to,
ending up with an ordered set $z'$ of entities which are answer candidates, very similar to traversing the KB. 
Furthermore, if the question requires more than one hop to answer, we can repeat the above procedure starting with $z'$.
% , which in the case of a real KB corresponds to taking a path of length 2. 
This is depicted pictorially in Figure~\ref{fig:my_label}.
% Although this idea is conceptually simple, a direct implementation would be prohibitively expensive.

We begin by first formalizing this idea in a probabilistic framework  in \S\ref{sec:reasoning}. 
In \S\ref{sec:computation}, we describe how the
expansion of entities
to mentions and the filtering of mentions
can be performed efficiently,
using sparse-matrix products
and MIPS algorithms \citep{johnson2017billion}.
Lastly we discuss a pretraining scheme for
constructing the mention representations in
\S\ref{sec:pretraining}.
% can be posed as a sparse matrix sparse vector multiplication operation
% and the filtering operation can be realized with a dual encoder approach utilizing efficient MIPS algorithms.
% However, a high performing dual encoder depends on learning high quality mention representations for which we describe a pretraining scheme by generating negative examples in Section \ref{sec:pretraining}. 

\textbf{Notation:}
We denote the given corpus as $\mathcal{D}=\{d_{1}, d_{2}, \ldots\}$, where each $d_k=(d_{k}^1, \ldots, d_{k}^{L_{k}})$ is a sequence of tokens.
% Our goal is to answer questions $q$, in natural language or semi-structured form with an entity from $\mathcal{D}$.
%, by extracting spans from the corpus.
%Specifically, we are interested in ``multi-hop'' questions, where $q$ implicitly involves following a path of relations starting from an initial set of entities and arriving at a final set of entities (the answer).
We start by running an entity linker over the corpus to identify mentions of a fixed set of entities $\mathcal{E}$.
Each mention $m$ is a tuple $(e_m, k_m, i_m, j_m)$ denoting that the text span $d_{k_m}^{i_m}, \ldots, d_{k_m}^{j_m}$ in document $k_m$ mentions the entity $e_m \in \mathcal{E}$, and the collection of all mentions in the corpus is denoted as $\mathcal{M}$.
Note that typically $|\mathcal{M}| \gg |\mathcal{E}|$.

\subsection{Differentiable Multi-Hop Reasoning}
\label{sec:reasoning}

We assume a \textit{weakly supervised} setting where during training we only know the final answer entities $a \in \mathcal{E}$ for a $T$-hop question.
We denote the latent sequence of entities which answer each of the intermediate hops as $z_0, z_1, \ldots, z_T \in \mathcal{E}$, where $z_0$ is mentioned in the question, and $z_T=a$.
We can recursively write the probability of an intermediate answer as:
\begin{equation} \label{eq:mhop-entity}
    \Pr(z_{t}|q) = \sum_{z_{t-1} \in \mathcal{E}} 
    \Pr(z_{t}|q, z_{t-1}) \Pr(z_{t-1}|q)
\end{equation}
Here $\Pr(z_0|q)$ is the output of an entity linking system over the question, and $\Pr(z_t|q, z_{t-1})$ corresponds to a single-hop model
which answers the $t$-th hop, \textit{given} the entity from the previous hop $z_{t-1}$, by following the appropriate relation.
%In general, this needs to be re-evaluated for each possible $z_{t-1}$, which
%can be prohibitively expensive.
%\bd{cite previous work in this context}
Eq.~\ref{eq:mhop-entity} models reasoning over a chain of latent entities, but when answering questions over a text corpus, we must reason over entity \textit{mentions}, rather than entities themselves.
Hence $\Pr(z_t|q, z_{t-1})$ needs to be aggregated over all mentions of $z_t$, which yields
% which we denote as
% $M_{z_t} = \{m \in \mathcal{M}: e_m=z_t\}$.
\begin{equation}
    \label{eq:entity-scores}
    \Pr(z_t |q) =
    \sum_{m \in \mathcal{M}} \; \sum_{z_{t-1}\in\mathcal{E}}
    \Pr(z_t|m) \Pr(m|q, z_{t-1}) \Pr(z_{t-1}|q) 
\end{equation}
The interesting term to model in the above equation is $Pr(m|q, z_{t-1})$, which represents the relevance of mention $m$ given the question and entity
$z_{t-1}$.
Following the analogy of a KB, we first expand the entity $z_{t-1}$ to co-occuring mentions $m$ and use a learned scoring function to find the relevance of these mentions. 
Formally, let $F(m)$ denote a TFIDF vector for the document containing $m$, $G(z_{t-1})$ be the TFIDF vector of the \textit{surface form} of the entity from the previous hop,
and $s_t(m, z, q)$ be a learnt scoring function (different for each hop).
%and let $\mathbbm{1}\left(h(m) > \epsilon\right)$ be an appropriate  vector of 0/1 indicators.  
Thus, we model $\Pr(m|q, z_{t-1})$ as
\begin{equation}
\label{eq:mention-scores}
    \Pr(m|q,z_{t-1}) \propto \underbrace{\mathbbm{1}\{G(z_{t-1}) \cdot F(m) > \epsilon\}}_{\text{expansion to co-occurring mentions}} \times \underbrace{s_t(m, z_{t-1}, q)}_{\text{relevance filtering}}
\end{equation}
%The first term in the product computes a dense retrieval score, and the second term \textit{filters} it based on whether that mention would be retrieved in a TFIDF search using the entity.
Another equivalent way to look at our model in Eq.~\ref{eq:mention-scores} is that the second term retrieves mentions of the correct \textit{type} requested by the question in the $t$-th hop, and the first term filters these based on co-occurrence with $z_{t-1}$.
When dealing with a large set of mentions $m$,
we will typically retain only the top-$K$ relevant mentions.
% according to the relevance score.
% Further we want to only retain top-$k$ mentions according to the relevance score to eliminate noisy tail scores.
We will show that this joint modelling of co-occurrence and relevance is important for good performance,
as was also observed by
\cite{seo2019real}.

The other term left in Eq.~\ref{eq:entity-scores} is $\Pr(z|m)$, which is $1$ if mention $m$ refers to the entity $z$ else $0$,
based on the entity linking system.
In general, to compute Eq.~\ref{eq:entity-scores} the mention scoring of Eq.~\ref{eq:mention-scores} needs to be evaluated for all latent entity and mention pairs, which is prohibitively expensive. 
However, by restricting $s_t$ to be an inner product
we can implement this efficiently
% However, our construction based on inner products
% allows an efficient implementation 
(\S\ref{sec:computation}).
% In Section~\ref{sec:computation} we show an efficient way for evaluation.

To highlight the differentiability of the proposed overall scheme, we can represent the computation in Eq.~\ref{eq:entity-scores} as matrix operations.
We pre-compute the TFIDF term for all entities and mentions into a \textit{sparse} matrix, which we denote as
$A_{E\to M}[e, m] = \mathbbm{1}\left(G(e) \cdot F(m) > \epsilon\right)$.
Then entity expansion to co-occuring mentions can be done using a sparse-matrix by sparse-vector multiplication between $A_{E\to M}$ and $z_{t-1}$.
For the relevance scores, let $\mathbb{T}_K(s_t(m, z_{t-1}, q))$ denote the top-$K$ relevant mentions encoded as a \textit{sparse} vector in $\mathbb{R}^{|\mathcal{M}|}$.
Finally, the aggregation of mentions to entities can be formulated as multiplication with another sparse-matrix $B_{M\to E}$, which encodes \textit{coreference}, i.e. mentions corresponding to the same entity.
Putting all these together, using $\odot$ to denote element-wise product, and defining $Z_{t} = [\Pr(z_{t}=e_1|q); \ldots; \Pr(z_{t}=e_{|\mathcal{E}|}|q)]$,
we can observe that for large $K$ (i.e., as $K\to |\mathcal{M}|$),  Eq.~\ref{eq:entity-scores} becomes equivalent to:
\begin{equation}
    \label{eq:follow}
    Z_t = \text{softmax} \left( \left[ Z_{t-1}^TA_{E\to M} \odot \mathbb{T}_K(s_t(m, z_{t-1}, q)) \right] B_{M\to E} \right).
\end{equation}
Note that \emph{every operation in above equation is differentiable and between sparse matrices and vectors}: we will discuss efficient
implementations in \S\ref{sec:computation}.
Further, the number of non-zero entries in $Z_t$ is bounded by $K$, since we filtered (the element-wise product in Eq.~\ref{eq:follow}) to top-$K$ relevant mentions among TFIDF based expansion and since each mention can only point to a single entity in $B_{M\to E}$.
This is important, as it prevents the number of entries in $Z_t$ from exploding across hops (which might happen if, for instance, we added the relevance and TFIDF scores instead).

We can view $Z_{t-1}, Z_{t}$ as weighted multisets of entities, and $s_t(m, z, q)$ as implicitly selecting mentions
which correspond to a relation $R$.
Then Eq.~\ref{eq:follow} becomes a differentiable implementation of $Z_t = Z_{t-1}.\text{follow}(R)$, i.e. mimicking the graph traversal in a traditional KB. We thus call Eq.~\ref{eq:follow} a \textit{textual follow operation}.

\paragraph{Training and Inference.}
The model is trained end-to-end by optimizing the cross-entropy loss between $Z_T$, the weighted set of entities after $T$ hops, and the ground truth answer set $A$.
We use a temperature coefficient $\lambda$ when computing the softmax in Eq,~\ref{eq:follow}
since the inner product scores of the top-$K$ retrieved mentions are typically high values, which would otherwise result in very peaked distributions of $Z_t$.
We also found that taking a \textit{maximum} over the mention set of an entity $M_{z_t}$ in Eq.~\ref{eq:entity-scores} works better than taking a sum. 
This corresponds to optimizing only over the most confident mention of each entity, which works for corpora like Wikipedia that 
do not have much redundancy.
A similar observation was made by \citet{min2019discrete} in weakly supervised settings.

\subsection{Efficient Implementation}
\label{sec:computation}

% From \eqref{eq:follow} one can see that DrKIT has four components: 1) Sparse TF-IDF mention encoding in form a sparse matrix $A_{M\to E}$, 2) efficient entity-mention expansion, 3) efficient top-$k$ mention relevance scoring, and 4) question and mention encoders. 
% Below we discuss tricks and details of each component needed for an efficient implementation of DrKIT.

\paragraph{Sparse TFIDF Mention Encoding.}
To compute the sparse-matrix $A_{E\to M}$ for entity-mention expansion in Eq.~\ref{eq:follow},
the TFIDF vectors $F(m)$ and $G(z_{t-1})$ are constructed over unigrams and bigrams, hashed to a vocabulary of $16M$ buckets.
While $F$ computes the vector from the whole passage around $m$, $G$ only uses the surface form of $z_{t-1}$.
This corresponds to retrieving all mentions in a document using $z_{t-1}$ as the query.
We limit the number of retrieved mentions per entity to a maximum of $\mu$,
which leads to  a
$|\mathcal{E}| \times |\mathcal{M}|$ sparse-matrix.
% with each row having no more than $\mu$ non-zero entries.

\begin{wrapfigure}{r}{0.4\textwidth}
\centering
\vspace{-6mm}
\includegraphics[width=\linewidth, trim={6mm 10mm 2mm 2mm}, clip]{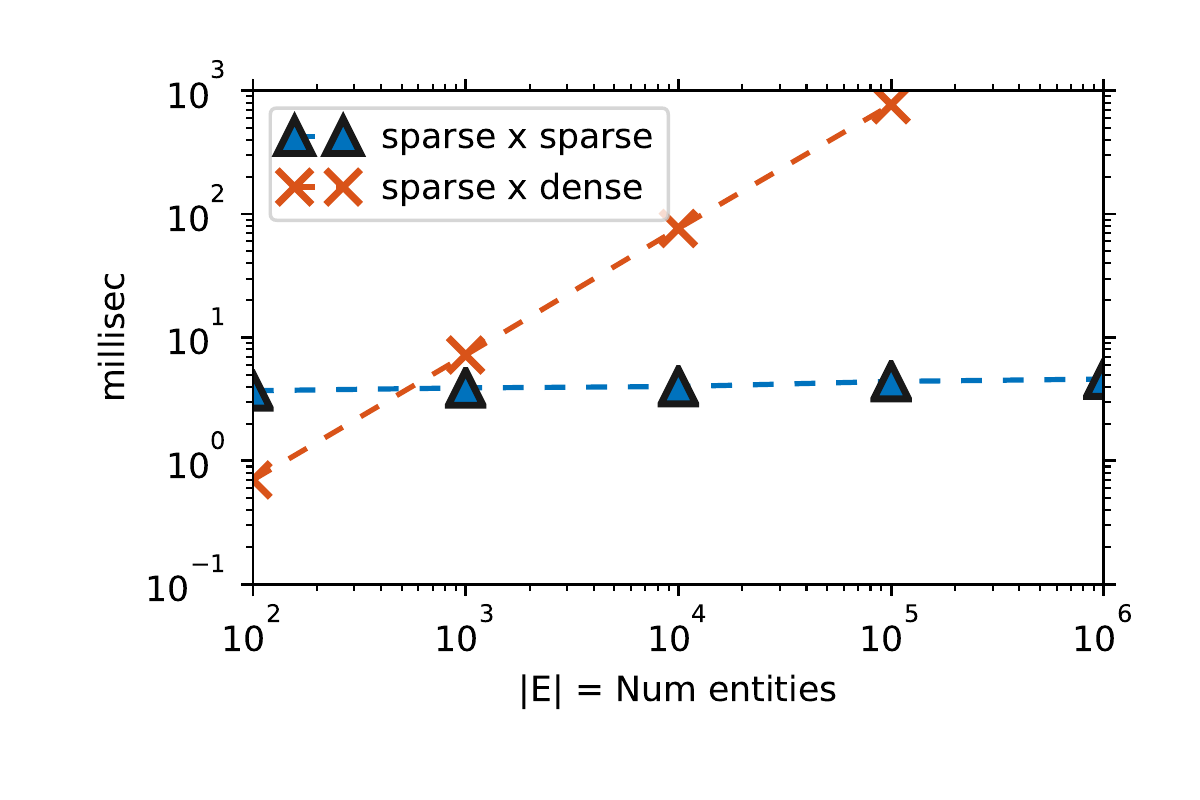}
%\vspace{-2mm}
\caption{\small Runtime on a single K80 GPU when using ragged representations for
implementing sparse-matrix vector product,
vs the default sparse-matrix times dense vector product available in TensorFlow.
$|\mathcal{E}| > 10^5$ leads to OOM for the latter.}
\label{fig:spspvmul}
\vspace{-3mm}
\end{wrapfigure}
\vspace{-2mm}
\paragraph{Efficient Entity-Mention expansion.}
% As mentioned previously,
The expansion from a set of entities to mentions
occurring around them can be computed using the sparse-matrix
by sparse-vector product
$Z_{t-1}^T A_{E\to M}$.
% The expansion from a set of entities $z$ to $m$ can be expressed as multiplication of a precomputed sparse matrix $A_{E\rightarrow M}$ with sparse (weighted) entity vector $z$.
A simple lower bound for multiplying a sparse $|\mathcal{E}| \times |\mathcal{M}|$ matrix, with maximum $\mu$ non-zeros in each row, by a sparse $|\mathcal{E}| \times 1$ vector with $K$ non-zeros is $\Omega(K\mu)$. 
% Since any sparse matrix-sparse vector multiplication algorithm has to access the nonzero entries in $k$ rows of $A_{E\rightarrow M}$ and each column of $A_{E\rightarrow M}$ has $\mu$ nonzeros.
Note that this lower bound is independent of the size of matrix $A_{E\rightarrow M}$, or in other words independent of the number of entities or mentions.
To attain the lower bound, the multiplication algorithm must be vector driven, because any matrix-driven algorithms need to at least iterate over all the rows.
% , e.g. when using DCSR and CSR formats.
Instead we \textit{slice} out the relevant rows from $A_{E\to M}$.
% To remain (almost) independent of of number of entities or mentions, our multiplication should be vector-driven and being able to slice out relevant rows from $A_{E\rightarrow M}$ efficiently.
To enable this our solution is to represent the sparse-matrix $A_{E\rightarrow M}$ as two row-wise \textit{lists
of variable-sized lists} of the indices
and values of the non-zero elements,
respectively.
This results in a ``ragged'' representation of the matrix \citep{tfragged} which can be easily sliced corresponding to the non-zero entries in the vector in $O(\log |\mathcal{E}|)$ time.
We are now left with $K$ sparse-vectors with at most $\mu$ non-zero elements in each. We can add these $K$ sparse-vectors weighted by corresponding values from the vector $Z_{t-1}^T$ in $O(K \max\{K,\mu\})$ time.
Moreover, such an implementation is feasible with deep learning frameworks such as TensorFlow.
%\footnote{\url{https://www.tensorflow.org/guide/ragged_tensors}}.
We tested the scalability of our approach by varying the number of entities for a fixed density of mentions $\mu$
(from Wikipedia).
Figure~\ref{fig:spspvmul} compares our approach to the default sparse-matrix times dense-vector product available in TensorFlow.

\paragraph{Efficient top-$K$ mention relevance filtering:}
%\paragraph{Multi-hop Dense Retrieval.}
%Since the query does not contain the intermediate entities
%$Z_{t-1}$,
% For multi-hop questions, we would like to take the output of one textual follow operation and make it the input of a downstream textual follow operation.  To do this, for we use entity embeddings
% $E \in \mathbb{R}^{|\mathcal{E}| \times p}$,
% to construct an average embedding of the set $Z_{t-1}$,
% as $Z_{t-1}^TE$, define \( g_t(q) \equiv \tilde{g}_t(q) + Z_{t-1}^T \), with \( \tilde{g}_t(q) \) 
% a question-dependent function defined below.
% To avoid a large number of parameters in the model,
% we compute the entity embeddings as an average over
% the word embeddings of the tokens in the entity's
% surface form.

To make computation of Eq.~\ref{eq:follow} feasible, we need an efficient way to get top-$K$ relevant mentions related to 
an entity in $z_{t-1}$ for a given question $q$, without enumerating all possibilities. 
A key insight is that by restricting the scoring function $s_t(m, z_{t-1}, q)$ to an inner product, we can easily approximate a parallel version of this computation, across all mentions $m$. %, and all possible pairs of entities $z_{t-1}, z_{t}$.  
To do this, let $f(m)$ be a dense encoding of $m$, and $g_t(q, z_{t-1})$ be a dense encoding of the question $q$ for the $t$-th hop, both in $\mathbb{R}^p$ (the details of the dense encoding is provided in next paragraph), then the scoring function $s_t(m, z_{t-1}, q)$ becomes
\begin{equation}
    s_t(m, z_{t-1}, q) \propto \exp\left\lbrace f(m) \cdot  g_t(q, z_{t-1}) \right\rbrace,
\end{equation}
which can be computed in parallel by multiplying a matrix $f(\mathcal{M}) = [f(m_1); f(m_2); \ldots]$ with $g_t(q, z_{t-1})$.  
Although this matrix will be very large for a realistic corpus,  since eventually we are only interested in the top-$K$ values, we can use an approximate algorithm for Maximum Inner Product Search (MIPS) \citep{andoni2015practical,shrivastava2014asymmetric} to find the $K$ top-scoring elements.
The complexity of this filtering step using MIPS is roughly $O(Kp \; \mathrm{polylog} |\mathcal{M}|)$.

\paragraph{Mention and Question Encoders.}
Mentions are encoded by passing the passages
they are contained in through a BERT-large \citep{devlin2018bert}
model (trained as described in \S\ref{sec:pretraining}). 
Suppose mention $m$ appears in passage $d$,
starting at position $i$ and ending at position $j$.
Then $f(m) = W^T[H^d_i; H^d_j]$, where $H^d$ is the
sequence of embeddings output from BERT,
and $W$ is a linear projection to size $p$.
The queries are encoded with a smaller BERT-like model: specifically, they are
tokenized with WordPieces \citep{schuster2012japanese}, appended to a special
\texttt{[CLS]} token, and then passed through a $4$-layer Transformer network
\citep{vaswani2017attention}
with the same architecture as BERT, producing an output sequence $H^q$.  The $g_t$ functions
are defined similarly to the BERT model used for SQuAD-style QA.
For each hop $t=1,\ldots,T$, we add two additional Transformer layers on top of $H^q$,
which will be trained to produce MIPS queries from the \texttt{[CLS]} encoding;
the first added layer produces a MIPS query $H^q_{st}$ to retrieve a start token, 
and the second added layer a MIPS query $H^q_{en}$ to retrieve an end token.
We concatenate the two and define $\tilde{g}_t(q) = V^T[H^q_{st}; H^q_{en}]$.
Finally, to condition on current progress we add the embeddings of $z_{t-1}$. Specifically, we use entity embeddings
$E \in \mathbb{R}^{|\mathcal{E}| \times p}$,
to construct an average embedding of the set $Z_{t-1}$,
as $Z_{t-1}^TE$, and define $ g_t(q, z_{t-1}) \equiv \tilde{g}_t(q) + Z_{t-1}^TE$.
%with \( \tilde{g}_t(q) \) 
% a question-dependent function defined below.
To avoid a large number of parameters in the model,
we compute the entity embeddings as an average over
the word embeddings of the tokens in the entity's
surface form. The computational cost of the question encoder $g_t(q)$ is $O(p^2)$.

Thus our total computational complexity to answer a query is $\tilde{O}(K \max\{K,\mu\} + Kp + p^2)$ (almost independent to number of entities or mentions!),
with $O(\mu|\mathcal{E}| + p|\mathcal{M}|)$ memory to store the pre-computed matrices and mention index.\footnote{Following standard convention, in $\tilde{O}$ notation we suppress $\text{poly}\log$ dependence terms.}
    
\subsection{Pretraining the Index}
\label{sec:pretraining}
% \vbc{Mention-Aware Pretraining? To distinguish from BERT}

Ideally, we would like to train the mention encoder
$f(m)$ end-to-end using labeled QA data only.
However, this poses a challenge
when combined with approximate nearest neighbor
search---since after every update to the parameters of $f$,
one would need to recompute the embeddings of all mentions
in $\mathcal{M}$.  We thus adopt a staged training approach: 
we first pre-train a mention encoder $f(m)$, then 
compute and index embeddings for all mentions once,
keeping these embeddings fixed when training the downstream QA task.
Empirically, we observed that 
using BERT representations ``out of the box''
do not capture the kind of information our task
requires (Appendix \ref{app:index-analysis}), 
%so how to
%perform this pretraining is an important research question. \vbc{rephrase} 
and thus, pretraining the encoder to capture better mention understanding is a crucial step.

One option adopted by previous researchers \citep{seo2018phrase}
is to fine-tune BERT on
SQuAD \citep{rajpurkar2016squad}.
However, SQuAD is limited to only $536$ articles from
Wikipedia, leading to a very specific distribution of questions,
and is not focused on entity- and relation-centric questions.  
Here we instead train the mention encoder using distant supervision from a KB.
% facts aligned with text to fine-tune BERT.

Specifically, assume we are given an open-domain KB
consisting of facts $(e_1, R, e_2)$ specifying that the relation
$R$ holds between the subject $e_1$ and the object $e_2$.
Then for a corpus of entity-linked text passages $\{d_k\}$,
we automatically identify tuples $(d, (e_1, R, e_2))$ such that
$d$ mentions both $e_1$ and $e_2$.
Using this data, we learn to answer slot-filling queries
in a reading comprehension setup,
where the query $q$ is
constructed from the surface form of the subject
entity $e_1$
and a natural language description of $R$
(e.g. ``Jerry Garcia, birth place, ?''),
and the answer $e_2$ needs to be extracted from the passage $d$.
Using string representations in $q$ ensures our pre-training
setup is similar to the downstream task.
In pretraining, we use the same scoring function as in previous section, but
over all \textit{spans} $m$ in the passage:
\begin{equation}
    s(m, e_1, q) \propto \exp\left\lbrace f(s) \cdot g(q, e_1)  \right\rbrace.
\end{equation}
Following \citet{seo2016bidirectional},
we normalize start and end probabilities of the span
separately.

For effective transfer to the full corpus setting,
we must also provide negative instances during pretraining,
i.e.~query and passage pairs where the answer is
\textit{not} contained in the passage.
%The structured nature of the task
%allows us to easily construct hard negatives.
We consider three types of hard negatives:
(1) \textit{shared-entity negatives}, which pair
a query $(e_1, R, ?)$ with a passage which mentions $e_1$
but not the correct tail answer;
(2) \textit{shared-relation negative}, which pair
a query $(e_1, R, ?)$ with a passage mentioning
two other entities $e'_1$ and $e'_2$ in the same relation
$R$; and
(3) \textit{random negatives}, which pair queries with
random passages from the corpus.

For the multi-hop slot-filling experiments below,
we used WikiData \citep{vrandevcic2014wikidata} as our KB,
Wikipedia as the corpus, and SLING \citep{ringgaard2017sling}
to identify entity mentions.
We restrict $d$ be from
the Wikipedia article of the subject entity
to reduce noise.
Overall we collected $950K$ pairs over $550K$ articles.
For the experiments with MetaQA, we supplemented this data
with the corpus and KB provided with MetaQA, and string matching
for entity linking.

%This could be further scaled up by sampling more
%Wikipedia articles,
%and also adapted to other domains,
%e.g. PubMed articles paired with UMLS concepts
%\citep{bio}.

\section{Experiments}

\subsection{METAQA: Multi-Hop Question Answering with Text}
\paragraph{Dataset.}
We first evaluate DrKIT on
the MetaQA benchmark for multi-hop question answering \citep{zhang2017variational}.
MetaQA
consists of around $400K$ questions ranging from
$1$ to $3$ hops constructed by sampling
relation paths from a movies KB \citep{miller2016key}
and converting
them to natural language using templates.
The questions cover $8$ relations and their inverses,
around $43K$ entities,
and are paired with a corpus consisting
of $18K$ Wikipedia passages about those entities.  The questions
are all designed to be answerable using either the KB or the corpus,
which makes it possible to compare the performance of 
our ``virtual KB'' QA system to a plausible upper bound system that has access to a complete KB.
We used the same version of the data as
\citet{sun2019pullnet}. Details of the implementation are in Appendix \ref{app:metqa-implementation}.

% \paragraph{Details.}
% We use $p=400$ dimensional embeddings for the mentions
% and queries, and $200$-dimensional embeddings each for the start and end positions.
% This results in an index of size $750$MB.
% When computing $A_{E\to M}$, the entity to mention
% co-occurrence matrix,
% we only retain mentions in the top $50$ paragraphs
% matched with an entity, to ensure sparsity.
% Further we initialize the first $4$ layers of
% the question encoder with
% the Transformer network from pre-training.
% For the first hop, we assign $Z_0$ as a $1$-hot
% vector for the least frequent entity detected in
% the question using an exact match.
% The number of nearest neighbors $K$ and
% the softmax temperature $\lambda$ were tuned on the dev set
% of each task, and we found $K=10000$ and $\lambda=4$ to work
% best.  As noted above, we pretrain the index on a combination
% of the MetaQA corpus, using the KB provided with MetaQA for
% distance data, and the Wikidata corpus described above. % \vbc{Below?}

\begin{table}
  \begin{minipage}[b]{0.5\linewidth}
    \centering
    \small
 \begin{tabular}{@{}lccc@{}}
 \toprule
 \multicolumn{4}{c}{\textbf{MetaQA}}                               \\ \midrule
 \textbf{Model}    & \textbf{1hop} & \textbf{2hop} & \textbf{3hop} \\ \midrule
 DrQA (ots)             & 0.553         & 0.325         & 0.197         \\ \midrule
 KVMem$\dagger$             & 0.762         & 0.070         & 0.195         \\
 GraftNet$\dagger$           & 0.825         & 0.362         & 0.402         \\
 PullNet$\dagger$            & 0.844         & 0.810         & 0.782         \\ \midrule
 DrKIT (e2e)       & 0.844         & 0.860         & \textbf{0.876}         \\
 DrKIT (strong sup.) & \textbf{0.845}         & \textbf{0.871}         & 0.871         \\ \bottomrule
 \end{tabular}
  \end{minipage}%
  \begin{minipage}[b]{0.5\linewidth}
    \centering
    \small
 \begin{tabular}{@{}lccc@{}}
 \toprule
 \multicolumn{4}{c}{\textbf{WikiData}}                            \\ \midrule
 \textbf{Model}   & \textbf{1hop} & \textbf{2hop} & \textbf{3hop} \\ \midrule
 DrQA (ots, cascade)            & 0.287         & 0.141         &    0.070           \\
 PIQA (ots, cascade)            & 0.240         & 0.118         & 0.064         \\ \midrule
 PIQA (pre, cascade)           & 0.670         & 0.369         & 0.182         \\
 DrKIT (pre, cascade) & 0.816         & 0.404         & 0.198         \\ \midrule
 DrKIT (e2e)      & \textbf{0.834}         & \textbf{0.469}         & \textbf{0.244}         \\
\enskip --BERT index & 0.643         & 0.294         & 0.165         \\ \bottomrule
 \end{tabular}
\end{minipage}
\caption{
\small
\textbf{(Left)} MetaQA and \textbf{(Right)} WikiData Hits @1
for $1$-$3$ hop sub-tasks. ots: off-the-shelf without re-training.
$\dagger$: obtained from \citet{sun2019pullnet}.
cascade: adapted to multi-hop setting by repeatedly applying Eq.~\ref{eq:entity-scores}.
pre: pre-trained on slot-filling.
e2e: end-to-end trained on single-hop and multi-hop queries.
\label{tab:metaqa}
}
\end{table}

\begin{figure}
    \centering
    \includegraphics[width=0.33\linewidth]{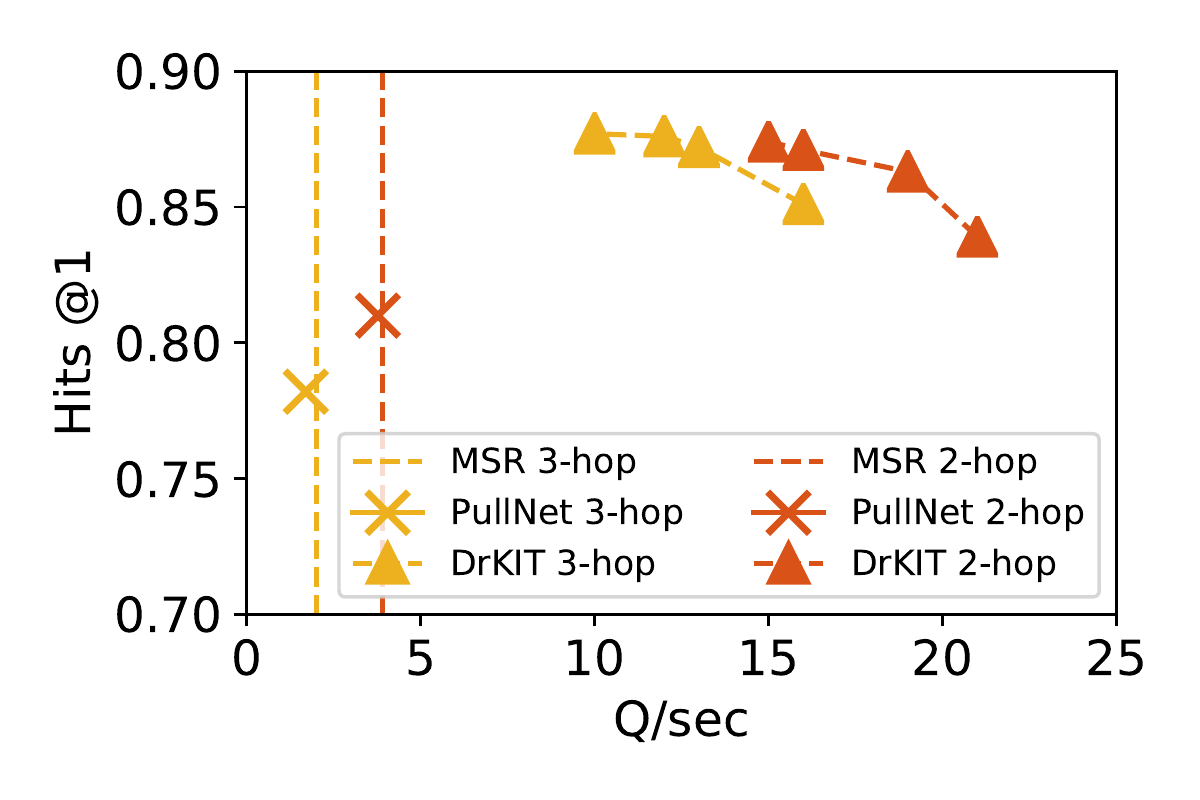}~
    \includegraphics[width=0.33\linewidth]{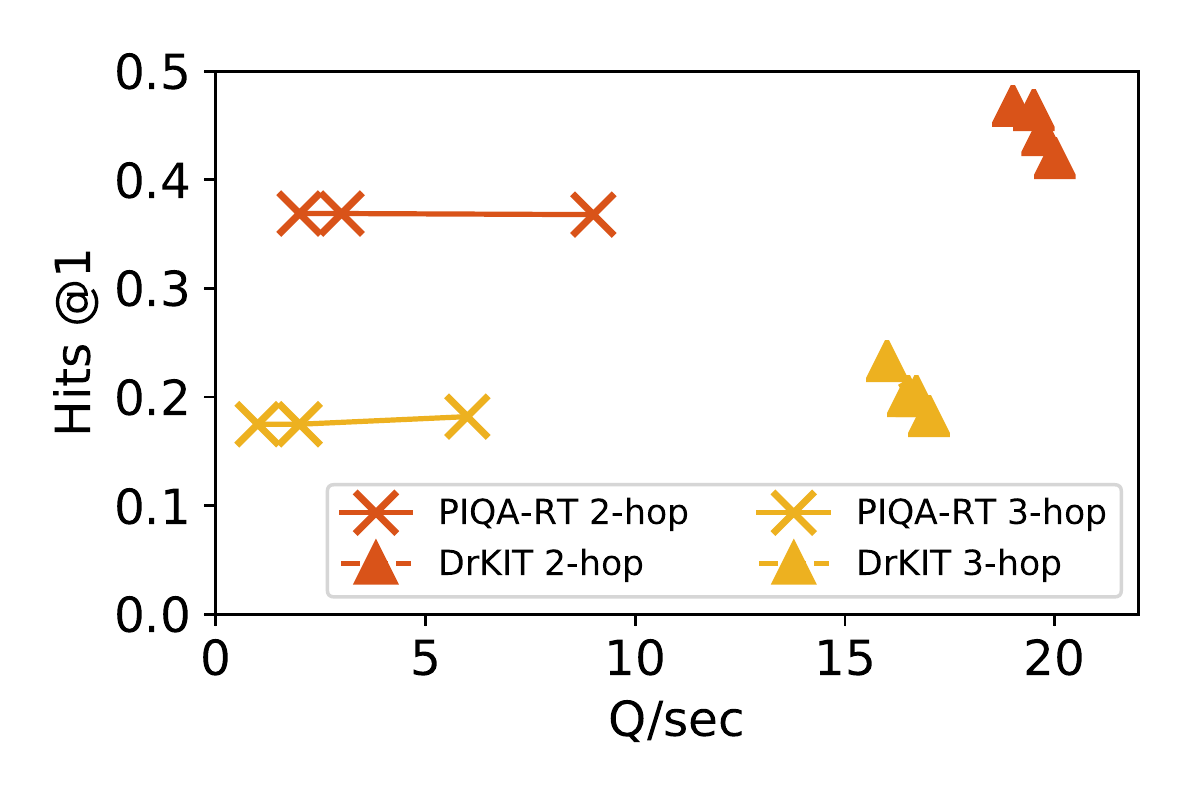}~
    \includegraphics[width=0.33\linewidth]{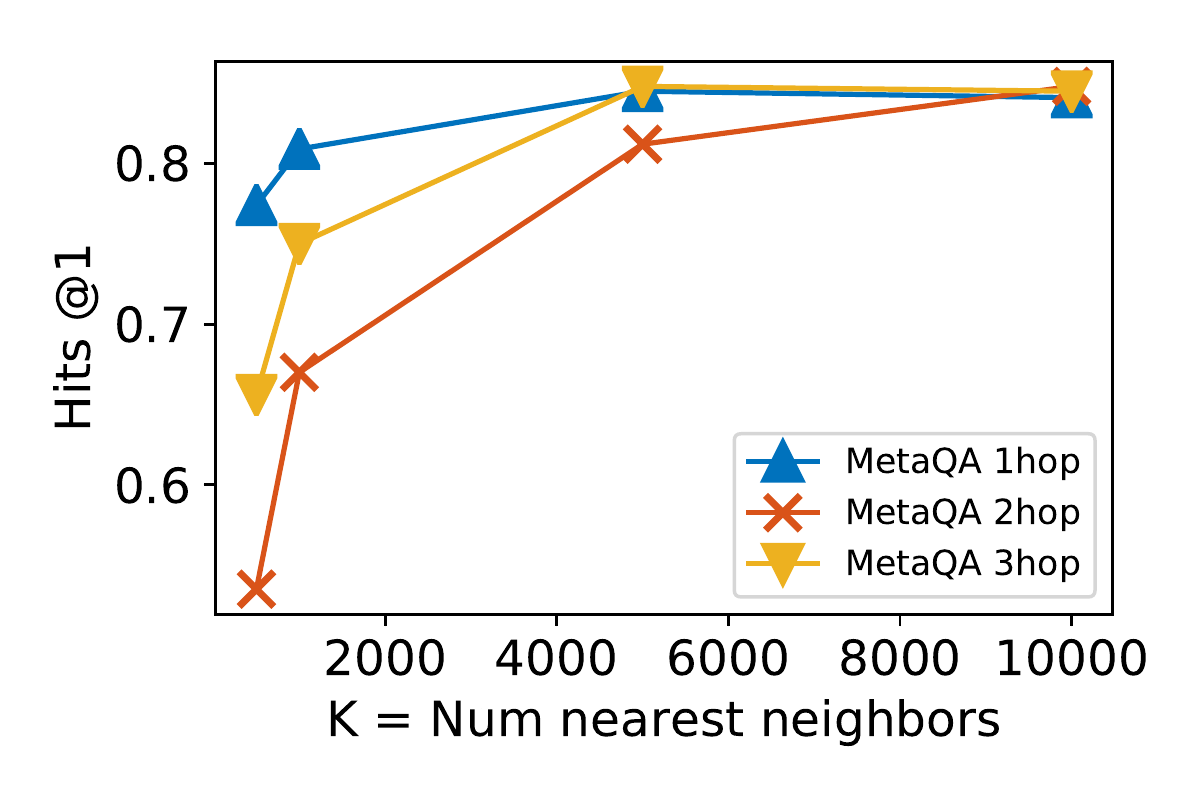}~
    \caption{ \small Hits @1 vs Queries/sec during inference on \textbf{(Left)} MetaQA and \textbf{(Middle)} WikiData tasks,
    measured on a single CPU server with $6$ cores.
    MSR: Multi-step Retriever model from \cite{das2019multi}
    (we only show Q/sec).
    \textbf{(Right)} Effect of varying number of nearest neighbors $K$
    during MIPS.
    }
    \label{fig:qps_plots}
\end{figure}

\paragraph{Results.}
Table~\ref{tab:metaqa} shows the accuracy of the top-most
retrieved entity (Hits@$1$) for the sub-tasks ranging from
$1$-$3$ hops, and compares to the state-of-the-art systems for the
text-only setting on these tasks.  DrKIT outperforms the prior
state-of-the-art by a large margin in the $2$-hop
and $3$-hop cases.  The strongest prior method,
PullNet \citep{sun2019pullnet, sun2018open}, uses a graph neural network model
with learned iterative retrieval from the corpus to answer multi-hop questions.
It uses
the MetaQA KB during training to identify shortest
paths between the question entity and answer entity,
which are used to supervise the text retrieval and reading modules.
DrKIT, on the other hand, has strong performance without such supervision, 
demonstrating its capability
for end-to-end learning.  (Adding the same intermediate supervision to DrKIT
does not even consistently improve performance---it gives DrKIT a small lift on 1- and 2-hop questions but does not help for 3-hop questions.)

DrKIT's architecture is driven, in part, by efficiency considerations: unlike PullNet, it is designed to answer questions with minimal processing at query time. Figure~\ref{fig:qps_plots} compares the tradeoffs between accuracy and inference time of DrKIT with PullNet as we vary $K$,
the number of dense nearest neighbors retrieved.
The runtime gains of DrKIT over PullNet range between $5$x-$15$x.

\paragraph{Analysis.}
%
% \begin{figure}
%   \begin{minipage}[b]{0.75\linewidth}
%     \centering
%     \small
% \begin{tabular}{@{}ccccccc@{}}
% \toprule
% \multirow{2}{*}{\textbf{KB Size}} & \multicolumn{2}{c}{\textbf{1hop}}          & \multicolumn{2}{c}{\textbf{2hop}}          & \multicolumn{2}{c}{\textbf{3hop}}          \\ \cmidrule(l){2-7} 
%                                   & \textbf{DrKIT} & \textit{\textbf{KB*}} & \textbf{DrKIT} & \textit{\textbf{KB*}} & \textbf{DrKIT} & \textit{\textbf{KB*}} \\ \midrule
% 25\% KB                           & 0.839          & \textit{--}               &                & \textit{--}               &                & \textit{--}               \\
% 50\% KB                           & 0.843          & \textit{0.680}             & 0.834          & \textit{0.521}            &                & \textit{0.597}            \\
% 100\% KB                          & 0.844          & \textit{0.970}             & 0.860           & \textit{0.999}            & 0.876          & \textit{0.914}            \\ \bottomrule
% \end{tabular}
%   \end{minipage}%
%   \begin{minipage}[b]{0.25\linewidth}
%     \centering
%     \includegraphics[width=\linewidth]{figures/K_effect.pdf}
%     \vspace{-1.5cm}
% \end{minipage}
% \caption{Effect of KB size used for pretraining the MetaQA index.
% * Best performance when using the same sized KB to directly answer
% questions at test time (c.f. \citep{sun2019pullnet}).
% \label{tab:kb-size}
% }
% \end{figure}
%
\begin{wraptable}{r}{0.42\linewidth}
\centering
\small
\vspace{-5mm}
\begin{tabular}{@{}lccc@{}}
\toprule
\textbf{Ablations}         & \textbf{1hop} & \textbf{2hop}  & \textbf{3hop}  \\ \midrule
DrKIT                      & 0.844         & 0.860           & 0.876          \\ \midrule
\enskip--Sum over $M_{z_t}$               & 0.837         &   0.823             & 0.797               \\
\enskip--$\lambda=1$ & 0.836         & 0.752          &     0.799           \\
\enskip--w/o TFIDF                & 0.845         &    0.548            &  0.488              \\
\enskip--BERT index               & 0.634         & 0.610           & 0.555          \\ \midrule
\multicolumn{4}{l}{\textit{Incomplete KB for pretraining}}                            \\ \midrule
25\% KB                    & 0.839         & 0.804               &  0.830              \\
50\% KB                    & 0.843         & 0.834           & 0.834          \\
\textit{(50\% KB-only)}    & \textit{0.680} & \textit{0.521} & \textit{0.597} \\ \bottomrule
\end{tabular}
\vspace{-2mm}

\end{wraptable}

We perform ablations on DrKIT for the
MetaQA data.
First, we empirically confirm that taking
a sum instead of max over the mentions of 
an entity hurts performance.
So does removing the softmax temperature
(by setting $\lambda=1$).
Removing the TFIDF component from Eq.~\ref{eq:mention-scores},
leads a large decrease in performance for
$2$-hop and $3$-hop questions.
This is because the TFIDF component
\textit{constrains}
the end-to-end learning to be along reasonable
paths of co-occurring mentions, preventing the search space from exploding.
The results also highlight the importance of the pretraining method of \S\ref{sec:pretraining},
as DrKIT over an index of BERT representations without pretraining
is $23$ points worse in the $3$-hop case.
% \footnote{
% Note, however, this is still better than applying a
% well trained single-hop model to the multi-hop case,
% as the results for DrQA and Graft-Net show.
% }
We also check the performance when the KB
used for pre-training is \textit{incomplete}.
Even with only $50\%$ edges retained,
we see good performance---better than PullNet
and the state-of-the-art for a KB-only method
(in \textit{italics}).

We analyzed 100 $2$-hop questions correctly answered by DrKIT
and found that for 83, the intermediate answers were also correct.
The other $17$ cases were all where the second hop asked about \textit{genre},
e.g. ``What are the genres of the films directed by Justin Simien?''.
We found that in these cases the intermediate answer was the same as the correct
final answer---essentially the model learned to answer the question in $1$ hop
and copy it over for the second hop.
Among incorrectly answered questions,
the intermediate accuracy was only $47\%$,
so the mistakes were evenly distributed across the two hops.

\subsection{WikiData: Multi-Hop Slot-Filling}

% \begin{figure}
% \begin{floatrow}
% \capbtabbox{%
% \begin{tabular}{@{}lccc@{}}
% \toprule
% \textbf{Model} & \textbf{1-hop} & \textbf{2-hop} & \textbf{3-hop} \\ \midrule
% DrQA$^\dagger$           & 0.287               &    0.141            &                \\
% PIQA$^\dagger$           & 0.240          & 0.118          & 0.064           \\
% PIQA-RT & 0.670          & 0.369          & 0.182               \\
% DrKIT     & 0.816          & 0.404          & 0.198          \\ \midrule
% % DrKIT (BERT index)     & 0.643          &  0.294         & 0.165          \\
% DrKIT (e2e)         & \textbf{0.834} & \textbf{0.469} & \textbf{0.244} \\ \bottomrule
% \end{tabular}
% }{%
% \caption{Hits@1 performance on WikiData slot-filling.
% The first four models are used in a \textit{cascaded} setting,
% where we repeatedly apply a single-hop trained model in the multi-hop case.
% $^\dagger$ Use templates for converting queries to natural language.
% \label{tab:WikiData}}
% }
% \ffigbox{%
% \includegraphics[width=\linewidth, trim={0.2cm 0.6cm 0 0}, clip]{figures/WikiData_qps.pdf}
% }{%
%   \caption{Hits @1 v Queries/sec on WikiData dataset, measured on a single $6$-score CPU machine.
%   Different points for DrKIT (PIQA-RT) correspond to different values
%   of $K$, the number of nearest neighbors retrieved in the dense part.}%
% }
% \end{floatrow}
% \end{figure}

The MetaQA dataset has been fairly well-studied, but has limitations
since it is constructed over a small KB.
% \vbc{What limitations?}
In this section we consider a new task,
% \vbc{some name?}
% \vbc{we should introduce details of the dataset and then benefits}
in a larger scale setting with many more relations,
entities and text passages.  The new dataset also lets us evaluate performance
in a setting where the test set contains documents and entities not seen at training time,
an important issue when devising a QA system that will be used in a real-world setting,
where the corpus and entities in the discourse change over time, and lets us perform analyses not possible with MetaQA, such as extrapolating from single-hop to multi-hop settings without retraining.
% \vbc{need to explain this limitation in MetaQA}, and evaluating performance on intermediate entities in a multi-hop reasoning chain.

\paragraph{Dataset.}
We sample two subsets of Wikipedia articles,
one for pre-training (\S\ref{sec:pretraining})
and end-to-end training,
and one for testing.
For each subset we consider the
set of WikiData entities mentioned in the articles,
and sample paths of $1$-$3$ hop relations among them,
ensuring that any intermediate entity has an in-degree of
no more than $100$.
Then we construct a semi-structured query by concatenating
the surface forms of the head entity with the path of relations
(e.g. ``Helene Gayle, employer, founded by, ?'').
The answer is the tail entity at the end of the path,
and the task is to extract it from the Wikipedia articles.
Existing slot-filling tasks \citep{levy2017zero,surdeanu2013overview}
focus on a \textit{single-hop}, \textit{static} corpus setting,
whereas our task considers a
\textit{dynamic} setting which requires the system to traverse the corpus.
For each setting,
we create a dataset with $10K$ articles,
$120K$ passages, $>$~$200K$ entities and 
$1.5M$ mentions,
resulting in an index of size about $2$gb.
We include example queries in Appendix \ref{app:datasets}.

% \begin{table}[!htbp]
% \centering
% \begin{tabular}{@{}lccc@{}}
% \toprule
% \textbf{Model} & \textbf{1-hop} & \textbf{2-hop}  \\ \midrule
% PIQA-rt           &    0.4377            &     0.2766               \\
% \end{tabular}
% \caption{Hits@1 performance on the Wikidata dataset.}
% \end{table}

% \begin{table}[!htbp]
% \centering
% \begin{tabular}{@{}lccc@{}}
% \toprule
% \textbf{Model} & \textbf{1-hop} & \textbf{2-hop}  \\ \midrule
% PIQA-rt           &    18.05            &     19.90               \\
% \end{tabular}
% \caption{It/sec speed performance on the Wikidata dataset.}
% \end{table}

\paragraph{Baselines.}
We adapt two publicly available
open-domain QA systems for this task --
DrQA\footnote{
\url{https://github.com/facebookresearch/DrQA}
} \citep{chen2017reading} and PIQA\footnote{
\url{https://github.com/uwnlp/denspi}
} \citep{seo2019real}.
While DrQA is relatively mature and widely used,
PIQA is recent, and similar to our setup
since it also
answers questions with minimal computation at query time.
It is broadly similar to a single textual follow operation in DrKIT,
but is not constructed to allow retrieved answers to be converted to entities and then used in 
subsequent processing, so it is not directly applicable to multi-hop queries.
We thus also consider a cascaded architecture which repeatedly applies
Eq.~\ref{eq:entity-scores}, using either of PIQA or DrQA
to compute $\text{Pr}(z_t|q, z_{t-1})$ against the corpus,
retaining at most $k$ intermediate answers in each step.
We tune $k$ in the range of $1$-$10$, since larger values make
the runtime infeasible.
Further, since these models were trained on natural language
questions, we use the templates released by \citet{levy2017zero}
to convert intermediate questions into natural text.\footnote{
For example, ``Helene Gayle. employer?'' becomes ``Who is the employer of Helene Gayle?''}
We test off-the-shelf versions of these systems,
as well as a version of PIQA re-trained on our our slot-filling 
data.\footnote{
We tuned several hyperparameters of PIQA on our data,
eventually picking the \textit{sparse first} strategy,
a sparse weight of $0.1$, and
a filter threshold of $0.2$. For the SQuAD trained version,
we also had to remove paragraphs smaller
than $50$ tokens since with these the model failed completely.}
We compare to a version of DrKIT trained only on single-hop
queries (\S\ref{sec:pretraining}) and similarly cascaded,
and one version trained end-to-end on the multi-hop queries.

\paragraph{Results.}
Table~\ref{tab:metaqa} (right) lists the Hits @1 performance on
this task.
Off-the-shelf open-domain QA systems perform poorly,
showing the challenging nature of the task.
Re-training PIQA on the slot-filling data
improves performance considerably,
but DrKIT trained on the same data improves on it.
A large improvement over these
cascaded architectures is seen with end-to-end training,
which is made possible by the differentiable operation
introduced in this paper.
We also list the performance of DrKIT when trained against an
index of fixed BERT-large mention representations.
While this is comparable to the re-trained version of PIQA,
it lags behind DrKIT pre-trained using the KB,
once again highlighting the importance of the scheme outlined
in \S\ref{sec:pretraining}.
We also plot the Hits @1 against Queries/sec for cascaded
versions of PIQA and DrKIT in Figure~\ref{fig:qps_plots} (middle).
We observe runtime gains of $2$x-$3$x to DrKIT due to the efficient implementation
of entity-mention expansion of \S\ref{sec:computation}.

\paragraph{Analysis.}
In order to understand where the accuracy gains for DrKIT come
from,
we conduct experiments on the dataset of slot-filling queries
released by \citet{levy2017zero}.
We construct an \textit{open} version of the task by collecting
Wikipedia articles of all subject entities in the data.
A detailed discussion is in Appendix~\ref{app:index-analysis}, and here we note the main findings.
PIQA trained on SQuAD only gets $30\%$ macro-avg accuracy
on this data, but this improves to $46\%$ when re-trained on
our slot-filling data.
Interestingly, a version of DrKIT which selects from \textit{all spans}
in the corpus performs similarly to PIQA ($50\%$), but
when using entity linking it significantly improves to $66\%$.
It also has $55\%$ accuracy in answering 
queries about \textit{rare} relations,
i.e. those observed $<$~$5$ times in its training data.
We also conduct probing experiments
comparing the representations learned using
slot-filling to those by vanilla BERT.
We found that while the two are comparable in
detecting \textit{fine-grained entity types},
the slot-filling version is significantly
better at encoding \textit{entity co-occurrence}
information.

\subsection{HotpotQA: Multi-Hop Information Retrieval}

\begin{table}[!htbp]
\small
\begin{minipage}[b]{0.6\linewidth}
\begin{tabular}{@{}lccccc@{}}
\toprule
\multirow{2}{*}{\textbf{Model}} & \multirow{2}{*}{\textbf{Q/s}}  & \multicolumn{4}{c}{\textbf{Accuracy}}                             \\ \cmidrule(l){3-6} 
                                &                                & \textbf{@2}    & \textbf{@5}    & \textbf{@10}   & \textbf{@20}   \\ \midrule
BM25$^\dagger$                            & --                             & 0.093          & 0.191          & 0.259          & 0.324          \\
PRF-Task$^\dagger$                         & --                             & 0.097          & 0.198          & 0.267          & 0.330          \\
BERT re-ranker$^\dagger$                   & --                             & 0.146          & 0.271          & 0.347          & 0.409          \\
Entity Centric IR$^\dagger$                & 0.32$^\ast$                           & 0.230          & 0.482          & 0.612          & 0.674          \\ \midrule
DrKIT (WikiData)                      & \multirow{3}{*}{\textbf{4.26}$^\ast$} & 0.355          & 0.588          & 0.671          & \textbf{0.710} \\
DrKIT (Hotpot)                      &                                & \textbf{0.385} & 0.595          & 0.663          & 0.703          \\
DrKIT (Combined)                    &                                & 0.383          & \textbf{0.603} & \textbf{0.672} & \textbf{0.710} \\ \bottomrule
\end{tabular}
\end{minipage}\enskip
\begin{minipage}[b]{0.3\linewidth}
\begin{tabular}{@{}lcc@{}}
\toprule
\textbf{Model}     & \textbf{EM}    & \textbf{F1}    \\ \midrule
Baseline$^\dagger$           & 0.288          & 0.381          \\
\enskip $+$EC IR$^\ddagger$            & 0.354          & 0.462          \\
\enskip $+$Golden Ret$^\diamond$ & \textbf{0.379} & \textbf{0.486} \\
\enskip $+$DrKIT$^\dagger$            & 0.357          & 0.466          \\ \bottomrule
\end{tabular}
\end{minipage}
\caption{\small
\textbf{(Left)} Retrieval performance on the HotpotQA benchmark dev set.
Q/s denotes the number of queries per second during inference on a single $16$-core CPU.
Accuracy @$k$ is the fraction where \textit{both} the correct
passages are retrieved in the top $k$.
$^\dagger$: Baselines obtained from \cite{das-etal-2019-multi}.
For DrKIT, we report the performance when the index is pretrained
using the WikiData KB alone, the HotpotQA training questions alone,
or using both.
$^\ast$: Measured on different machines with similar specs.
\textbf{(Right)} Overall performance on the HotpotQA task,
when passing $10$ retrieved passages to a downstream reading comprehension
model \citep{yang2018hotpotqa}.
$^\ddagger$: From \cite{das-etal-2019-multi}.
$^\diamond$: From \cite{qi2019answering}.
$^\dagger$: Results on the dev set.
\label{tab:hotpot}
}
\end{table}

% \begin{wraptable}{r}{0.38\linewidth}
% \centering
% \small
% \begin{tabular}{@{}lcc@{}}
% \toprule
% \textbf{Model}     & \textbf{EM}    & \textbf{F1}    \\ \midrule
% Baseline$^\dagger$~$^\ast$           & 0.288          & 0.381          \\
% \enskip $+$EC IR$^\ddagger$            & 0.354          & 0.462          \\
% \enskip $+$Golden Retriever$^\diamond$ & \textbf{0.379} & \textbf{0.486} \\
% \enskip $+$DrKIT$^\ast$            & 0.357          & 0.466          \\ \bottomrule
% \end{tabular}
% \caption{\small 
% }
% \end{wraptable}

\paragraph{Dataset.}
HotpotQA \citep{yang2018hotpotqa} is a recent dataset of over $100$K crowd-sourced
multi-hop questions and answers over introductory Wikipedia passages.
We focus on the open-domain \textit{fullwiki} setting where the two gold
passages required to answer the question are not known in advance.
The answers are free-form spans of text in the passages,
not necessarily entities,
and hence our model which selects entities is not directly applicable here.
Instead, inspired by
recent works \citep{das-etal-2019-multi,qi2019answering},
we look at the challenging sub-task of \textit{retrieving} the passages required
to answer the questions from a pool of $5.23$M.
This is a multi-hop IR task, since for many questions
at least one passage may be $1$-$2$ hops away from the entities in the question.
Further, each passage is about an entity (the title entity of that Wikipedia page),
and hence retrieving passages is the same as identifying the title entities of
those passages.
We apply DrKIT to this task of identifying the two entities for each question,
whose passages contain the information needed to answer that question.
Then we pass the top $10$ passages identified this way to a standard reading
comprehension architecture from \cite{yang2018hotpotqa} to select
the answer span.

\paragraph{Setup.}
We use the Wikipedia abstracts released by \cite{yang2018hotpotqa} as the
text corpus.\footnote{
\url{https://hotpotqa.github.io/wiki-readme.html}}
The total number of entities is the same as the number of abstracts,
$5.23$M,
and we consider hyperlinks in the text as mentions of the entities to whose
pages they point to,
leading to $22.8$M total mentions in an index of size
$34$GB.
For pretraining the mention representations,
we compare using the WikiData KB as described in \S \ref{sec:pretraining}
to directly using the HotpotQA training questions,
with TFIDF based retrieved passages as negative examples.
We set $A_{E\to M}[e,m]=1$ if either the entity $e$ is mentioned
on the page of the entity denoted by $m$, or vice versa.
For entity linking over the questions,
we retrieve the top $20$ entities based on the match between
a bigram based TFIDF vector of the question with a similar vector derived from
the surface form of the entity (same as the title of the Wiki article).
We found that the gold entities that need to be retrieved are within
$2$ hops of the entities linked in this manner for $87\%$ of the dev examples.

Unlike the MetaQA and WikiData datasets, however, for HotpotQA we do
not know the number of hops required for each question in advance.
Instead, we run DrKIT for $2$ hops for each question,
and then take a weighted average of the distribution over entities
after each hop $Z^\ast = \pi_0 Z_0 + \pi_1 Z_1 + \pi_2 Z_2$.
$Z_0$ consists of the entities linked to the question itself,
rescored based on an encoding of the question,
since in some cases one or both the entities to be retrieved
are in this set.\footnote{
For example, for the question ``How are elephants connected to Gajabrishta?'',
one of the passages to be retrieved is ``Gajabrishta'' itself.}
$Z_1$ and $Z_2$ are given by Eq.~\ref{eq:follow}.
The mixing weights $\pi_i$ are the softmax outputs of a classifier
on top of another encoding of the question,
learnt \textit{end-to-end} on the retrieval task.
This process can be viewed as soft mixing of different templates
ranging from $0$ to $2$ hops for answering a question,
similar to NQL \citep{cohen2019neural}.

\paragraph{Results.}
We compare our retrieval results to those presented in
\cite{das-etal-2019-multi} in Table~\ref{tab:hotpot} (Left).
We measure the accuracy @$k$ retrievals,
which is the fraction of questions for which
\textit{both} the required passages are
in the top $k$ retrieved ones.
We see an improvement in accuracy across the board,
with much higher gains @$2$ and @$5$.
The main baseline is the entity-centric IR approach
which runs a BERT-based re-ranker on $200$ pairs
of passages for each question.
Importantly, DrKIT also improves by over $10$x in terms
of queries per second during inference.
Note that the inference time is measured using a batch
size of $1$ for both models for fair comparison.
DrKIT can be easily run with batch sizes up to $40$,
but the entity centric IR baseline cannot due to the large
number of runs of BERT for each query.
When comparing different datasets for pretraining the index,
there is not much difference between using the WikiData KB,
or the HotpotQA questions. The latter has a better accuracy @$2$,
but overall the best performance is when using a combination of both.

In Table~\ref{tab:hotpot} (Right),
we check the performance of the baseline reading comprehension
model from \cite{yang2018hotpotqa}, when given the passages
retrieved by DrKIT.
While there is a significant improvement over the baseline which
uses a TFIDF based retrieval,
we see only a small improvement over the passages retrieved
by the entity-centric IR baseline,
despite the significantly improved accuracy @$10$ of DrKIT.
Among the $33\%$ questions where the top $10$ passages
do not contain both the correct passages,
for around $20\%$ the passage containing the answer
is also missing.
We conjecture this percentage is lower for the entity-centric
IR baseline, and the downstream model is able to answer
some of these questions without the other supporting passage.

\begin{table}[!tbp]
\setlength\tabcolsep{5pt}
\begin{tabular}{@{}lcccccccc@{}}
\toprule
\multirow{2}{*}{\textbf{System}} & \multicolumn{2}{c}{\textbf{Runtime}} & \multicolumn{2}{c}{\textbf{Answer}} & \multicolumn{2}{c}{\textbf{Sup Fact}} & \multicolumn{2}{c}{\textbf{Joint}} \\ \cmidrule(l){2-3} \cmidrule(l){4-5} \cmidrule(l){6-7} \cmidrule(l){8-9}
                               & \textbf{\#Bert}  &  \textbf{s/Q}        & \textbf{EM}      & \textbf{F1}      & \textbf{EM}       & \textbf{F1}       & \textbf{EM}      & \textbf{F1}     \\ \midrule
Baseline \citep{yang2018hotpotqa}   & --       &      --        & 25.23            & 34.40            & 5.07              & 40.69             & 2.63             & 17.85           \\
% Cog Graph  \citep{ding2019cognitive} &  &               & 37.12            & 48.87            & 22.82             & 57.69             & 12.42            & 34.92           \\
Golden Ret   \citep{qi2019answering} & --  &   1.4$^\dagger$       & 37.92            & 48.58            & 30.69             & 64.24             & 18.04            & 39.13           \\
Semantic Ret \citep{nie2019revealing}&  50$^\ast$ &   40.0$^\ddagger$        & 45.32            & 57.34            & 38.67             & 70.83             & 25.14            & 47.60           \\
HGN \citep{fang2019hierarchical}  &   50$^\ast$  & 40.0$^\ddagger$  & 56.71            & 69.16            & \textbf{49.97}    & 76.39             & \textbf{35.63}   & 59.86           \\
Rec Ret  \citep{asai2020learning} & 500$^\ast$ &  133.2$^\dagger$  & \textbf{60.04}   & \textbf{72.96}   & 49.08             & \textbf{76.41}    & 35.35            & \textbf{61.18}  \\ \midrule
DrKIT + BERT  & \textbf{1.2}$^\diamond$ &  \textbf{1.3}        & 42.13            & 51.72            & 37.05             & 59.84             & 24.69            & 42.88           \\ \bottomrule
\end{tabular}
\caption{\small
Official leaderboard evaluation on the test set of HotpotQA.
\#Bert refers to the number of calls to BERT \citep{devlin2018bert} in the model.
s/Q denotes seconds per query (using batch size $1$) for inference on a single $16$-core CPU.
Answer, Sup Fact and Joint are the official evaluation metrics for HotpotQA.
$^\ast$: This is the minimum number of BERT calls based on model and hyperparameter
descriptions in the respective papers.
$^\dagger$: Computed using code released by authors, using a batch size of $1$.
$^\ddagger$: Estimated based on the number of BERT calls, using $0.8$s as the time
for one call (without considering overhead due to other computation in the model).
$^\diamond$: One call to a $5$-layer Transformer, and one call to BERT.
\label{tab:hotpot_test}}
\end{table}

Lastly, we feed the top $5$
passages retrieved by DrKIT to an improved
answer span extraction model based on BERT.
This model implements a standard architecture for
extracting answers from text,
and is trained to predict both the answers and the supporting facts.
Details are included in Appendix~\ref{app:hotpot_answer}.
Table~\ref{tab:hotpot_test} shows the performance of this
system on the HotpotQA test set, compared with other
recently published models on the leaderboard.\footnote{
As of February 23, 2020:
\url{https://hotpotqa.github.io/}.}
In terms of accuracy, DrKIT+BERT reaches a modest score of
$42.88$ joint F1,
but is considerably faster (up to $100$x)
than the models which outperform it.

% \vspace{-1mm}
\section{Related Work}
% \vspace{-1mm}

% \gn{Haven't read this carefully, but maybe put it after the experiments.}

Neural Query Language (NQL) \citep{cohen2019neural} defines differentiable templates
for multi-step access to a symbolic KB,
in which relations between entities are \textit{explicitly}
enumerated.
Here, we focus on the case where the 
relations are implicit in mention representations derived from text.
% The key idea is to encode sets of entities as
% $k$-hot vectors and implement relation-following as sparse matrix multiplications.
% % This work develops similar machinery, but extends it to multi-step access to information in a
% % \textit{text corpus}, a more challenging problem.
% We also represent sets of entities as $k$-hot vectors,
% but implement relation-following as a nearest neighbor search over the indexed
% representations of text mentions.
Knowledge Graph embeddings 
\citep{bordes2013translating,yang2014embedding,dettmers2018convolutional}
attach continuous representations to discrete symbols which allow
them to be incorporated in deep networks \citep{yang2019leveraging}.
Embeddings often allow generalization to unseen
facts using relation patterns,
but text corpora are more complete
in the information they contain.

\citet{talmor18compwebq} also examined answering compositional
questions by treating a text corpus (in their case the entire web) as a KB.
However their approach consists of parsing the query into
a computation tree,
and running a black-box QA model on its leaves separately,
which \textit{cannot} be trained end-to-end.
%Their approach consisted of two stages -- a semantic parser for decomposing
%the query into a computation tree, and a black-box QA model for answering
%simple (1-hop) questions at the leaf nodes of the computation tree.
%Since these two components interact symbolically (the decomposed questions from
%the semantic parser form the input to the QA model), end-to-end learning from
%complex QA pairs is not possible. Instead the semantic parser was
%trained on noisy supervision generated heuristically.
% Our approach relies on pre-training stage for learning the contextual
% representations in the corpus index. Following that a complex QA
% model can be learned end-to-end using only the answers as supervision.
Recent papers have also looked at complex QA using graph neural networks
\citep{sun2018open,cao2019bag,xiao2019dynamically}
or by identifying paths of entities in
text \citep{jiang2019explore,kundu2018exploiting,dhingra2018neural}.
These approaches rely on identifying a small relevant pool
of evidence documents containing the information required for multi-step QA.
Hence, \citet{sun2019pullnet} and \citet{ding2019cognitive},
incorporate a dynamic retrieval process to add text about entities
identified as relevant in the previous layer of the model.
% These methods show impressive performance on multi-hop QA, but the
% reasoning process employed is opaque
% since intermediate representations in the network are uninterpretable.
Since the evidence text is processed in a query-dependent manner,
the inference speed is slower than when it is pre-processed into an
indexed representation (see Figure~\ref{fig:qps_plots}).
The same limitation is shared by methods which perform multi-step
retrieval interleaved with a reading comprehension model
\citep{das2019multi,feldman2019multi,lee-chang-toutanova:2019:ACL2019}.
% Our approach explicitly produces sets of intermediate answers
% for a complex questions, which we evaluate in \S \ref{sec:intermediate}.

% Phrase-indexed QA (PIQA)
% \citep{seo2018phrase,seo2019real}
% uses MIPS operations, combined with sparse retrieval,
% for \textit{open-domain} question answering against a corpus
% index.
% This means that the index must store a potentially unbounded
% amount of information, since in principle any query might be
% posed against it.
% Here, we restrict the index to only slot-filling
% queries, and defer more complex queries to a separate reasoning
% module.
% The restricted setting allows us to construct harder negatives
% for pre-training the index, which leads to improved performance
% on slot-filling compared to PIQA.   The focus on slot-filling queries, which can be naturally combined compositionally, also allows us to extend the system to answer multi-hop queries.

% \vspace{-1mm}
\section{Conclusion}
% \vspace{-1mm}

We present DrKIT, a differentiable module that is capable of answering multi-hop questions directly using a large entity-linked text corpus. 
DrKIT is designed to imitate traversal in KB over the text corpus,
providing ability to follow relations in the ``virtual'' KB over text. 
We achieve state-of-the-art results on the MetaQA dataset for answering natural language questions,
with a $9$ point increase in the $3$-hop case.
We also developed an efficient implementation using sparse operations and inner product search, which led to a $10$-$100$x increase in Queries/sec over baseline approaches.

\subsubsection*{Acknowledgments}
Bhuwan Dhingra was supported by a Siemens
fellowship during this project.
This work was supported in part by ONR Grant N000141812861, Google,
Apple, and grants from NVIDIA.

\bibliography{iclr2020_conference}
\bibliographystyle{iclr2020_conference}

\appendix

\section{METAQA: Implementation Details}
\label{app:metqa-implementation}
We use $p=400$ dimensional embeddings for the mentions
and queries, and $200$-dimensional embeddings each for the start and end positions.
This results in an index of size $750$MB.
When computing $A_{E\to M}$, the entity to mention
co-occurrence matrix,
we only retain mentions in the top $50$ paragraphs
matched with an entity, to ensure sparsity.
Further we initialize the first $4$ layers of
the question encoder with
the Transformer network from pre-training.
For the first hop, we assign $Z_0$ as a $1$-hot
vector for the least frequent entity detected in
the question using an exact match.
The number of nearest neighbors $K$ and
the softmax temperature $\lambda$ were tuned on the dev set
of each task, and we found $K=10000$ and $\lambda=4$ to work
best. We pretrain the index on a combination
of the MetaQA corpus, using the KB provided with MetaQA for
distance data, and the WikiData corpus.

\section{WikiData Dataset Statistics}
\label{app:datasets}

\begin{table}[!htbp]
\small
\centering
\begin{tabular}{@{}cccccccl@{}}
\toprule
\textbf{Task} & \textbf{\#train} & \textbf{\#dev} & \textbf{\#test} & \textbf{$|\mathcal{E}_{test}|$} & \textbf{$|\mathcal{M}_{test}|$} & \textbf{$|\mathcal{D}_{test}|$} & \textbf{Example}                                                                                                                                                                    \\ \midrule
1hop          & 16901            & 2467           & 10000           &     216K             & 1.2M             & 120K             & \begin{tabular}[c]{@{}l@{}}Q. Mendix, industry?\\ A. Enterprise Software\end{tabular}                                                                                               \\ \midrule
2hop          &   163607               &    398            &      9897           &      342K            &     1.9M             &      120K            & \begin{tabular}[c]{@{}l@{}}Q. 2000 Hel van het Mergelland, winner,\\ place of birth?\\ A. Bert Grabsch $\to$ Lutherstadt Wittenberg\end{tabular}                       \\ \midrule
3hop          &    36061              &     453           &       9899          &      261K            &          1.8M        &  120K                & \begin{tabular}[c]{@{}l@{}}Q. Magnificent!, record label, founded by,\\ date of death?\\ A. Prestige $\to$ Bob Weinstock $\to$\\ 14 Jan 2006\end{tabular} \\ \bottomrule
\end{tabular}
\caption{WikiData dataset \label{tab:wikidata-data}}
\end{table}

Details of the collected WikiData dataset are shown in
Table~\ref{tab:wikidata-data}.

\section{Index Analysis}
\label{app:index-analysis}

\begin{figure}
  \begin{minipage}[b]{0.60\linewidth}
    \centering
    \small
\begin{tabular}{@{}clcc@{}}
\toprule
\textbf{Probing}\\\textbf{Task}                                     & \textbf{Negative Example}                                                                             & \textbf{BERT} & \textbf{DrKIT} \\ \midrule
\begin{tabular}[c]{@{}c@{}}Shared\\ Entity\end{tabular}   & \begin{tabular}[c]{@{}l@{}} Neil Herron played for\\ West of Scotland.\end{tabular}  & 0.850         & 0.876          \\ \midrule
\begin{tabular}[c]{@{}c@{}}Shared\\ Relation\end{tabular} & \begin{tabular}[c]{@{}l@{}} William Paston was a\\ British politician.\end{tabular} & 0.715         & 0.846          \\ \bottomrule
\end{tabular}
  \end{minipage}%
  \begin{minipage}[b]{0.40\linewidth}
    \centering
    \includegraphics[width=\linewidth]{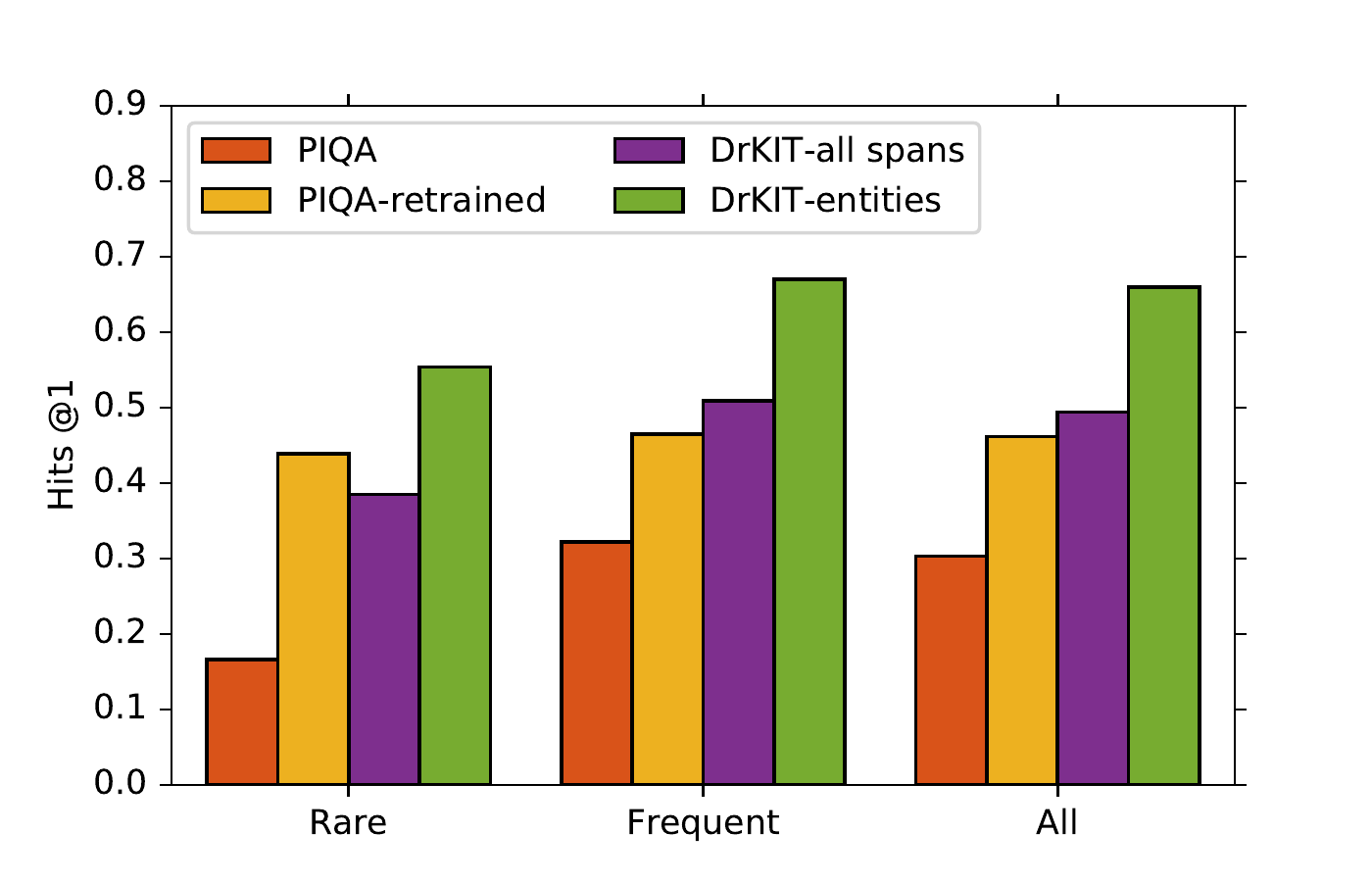}
    \vspace{-2cm}
\end{minipage}
\caption{ \textbf{Left: }F1 scores on Shared Entity and Shared Relation negatives. The negative examples are for the Query : (Neil Herron, occupation, ?).
\textbf{Right: }Macro-avg accuracy on the \citet{levy2017zero} relation extraction dataset.
    We split the results based on frequency of the relations in our WikiData training data.
    DrKIT-all spans refers to a variant of our model which selects from all spans
    in the corpus, instead of only entity-linked mentions.
\label{tab:analysis}
}
\end{figure}

\paragraph{Single-hop questions and relation extraction.} 
\citet{levy2017zero} released a dataset of $1M$ slot-filling
queries of the form $(e_1, R, ?)$
paired with Wikipedia sentences mentioning $e_1$, which
was used for training systems that answered single-step slot-filling
questions based on a small set of candidate passages.
Here we consider an \textit{open} version of the same task,
where answers to the queries must be extracted from a corpus rather
than provided candidates.
We construct the corpus by collecting and entity-linking all paragraphs in the
Wikipedia articles of all $8K$ subject
entities in the dev and test sets,
leading to a total of $109K$ passages.
After constructing the TFIDF $A_{E\to M}$
and coreference $B_{M\to E}$ matrices for this corpus,
we directly use our pre-trained index to answer
the test set queries.

Figure \ref{tab:analysis} (Right) shows the Hits@1 performance of the \cite{levy2017zero} slot-filling dataset. We report results on 2 subsets of relations in addition to all relations. The Rare subset comprises of relations with frequencies $<$ 5 in the training data while the 'Frequent' subset contains the rest. DrKIT on entity-mentions consistently outperforms the other phrase-based models showing the benefit of indexing only entity-mentions in single-hop questions over all spans.
Note that DrKit-entities has a high Hits@1 performance on the Rare relations subset, showing that there is generalization to less frequent data due to the natural language representations of entities and relations.

\paragraph{Probing Experiments}

Finally, to compare the representations learned by the BERT model fine-tuned on the WikiData slot-filling task, we design two probing experiments. In each experiment, we keep the parameters of the BERT model (mention encoders) being probed fixed and only train the query encoders. Similar to \citet{tenney2018what},
we use a weighted average of the layers of BERT here
rather than only the top-most layer, where the
weights are learned on the probing task.

In the first experiment, we train and test on shared-entity negatives. Good performance here means
the BERT model being probed encodes fine-grained
\textit{entity-type} information reliably%
\footnote{A reasonable heuristic for solving this
task is to simply detect an entity with the correct
type in the given sentence, since all sentences
contain the subject entity.}.  As shown in Table~\ref{tab:analysis},
BERT performs well on this task, suggesting it encodes fine-grained types well.

In the second experiment, we train and test only on
shared-relation negatives.  Good performance here means that the BERT model encodes \textit{entity co-occurrence} information reliably.  In this probe task, we see a large performance drop for BERT, suggesting it does not encode entity co-occurrence information well. The good performance of the DrKIT model on both  experiments suggests that fine-tuning on the slot-filling task primarily helps the contextual representations to also encode entity co-occurrence information, in addition to entity type information.

\section{HotpotQA Answer Extraction}
\label{app:hotpot_answer}

On HotpotQA, we use DrKIT to identify the top passages which are likely
to contain the answer to a question.
We then train a separate model to extract the answer from a concatenation
of these passages.
This model is a standard BERT-based architecture used for SQuAD
(see \cite{devlin2018bert} for details),
with a few modifications.
First, to handle boolean questions,
we train a $3$-way classifier on top of the \texttt{[CLS]}
representation from BERT to decide whether the question has
a ``span'', ``yes'' or ``no'' answer, respectively.
During inference, if this classifier has the highest probability
on ``span'' we extract a start and end position similar to \cite{devlin2018bert},
else we directly answer as ``yes'' or ``no''.

Second, to handle supporting fact prediction,
we prepend each sentence in the concatenated passages passed to BERT
with a special symbol \texttt{[unused0]},
and train a binary classifier on top of the representation of
each of these symbols output from BERT.
The binary classifier is trained to predict $1$ for sentences
which are supporting facts and $0$ for sentences which are not.
During inference, we take all sentences for which the output
probability of this classifier is $>0.5$ as supporting facts.

The training loss is an average of the loss for
the $3$-way classifier ($\mathcal{L}_{cls}$),
the sum of the losses for the supporting fact classifiers ($\mathcal{L}_{sp}$),
and the losses for the start and end positions of span answers
($\mathcal{L}_{st}$, $\mathcal{L}_{en}$):
\begin{equation}
    \mathcal{L} = (\mathcal{L}_{cls} + \mathcal{L}_{sp} + \mathcal{L}_{st} + \mathcal{L}_{en}) / 4
\end{equation}
We train the system on $5$ passages per question, provided in the distractor
setting of HotpotQA---$2$ gold ones and $3$ negatives from a TFIDF retriever.
We keep the gold passages at the beginning for $60\%$ of the examples,
and randomly shuffle all passages for the rest, since during inference
the correct passages are likely to be retrieved at the top by DrKIT.
Other hyperparameters include---batch size $32$, learning rate $5\times 10^{-5}$,
number of training epochs $5$, and a maximum combined passage length $512$.

\end{document}